\documentclass[sigconf]{acmart}

\usepackage{booktabs} 
\usepackage{algorithm}
\usepackage[noend]{algorithmic}
\usepackage{graphicx}
\usepackage{url}

\setcopyright{none}

\renewcommand\footnotetextcopyrightpermission[1]{}
\begin{document}

\title{Knowledge Base Completion using Web-Based Question Answering and Multimodal Fusion}

\author{Yang Peng}
\email{yangpengsnf@ufl.edu}
\affiliation{University of Florida}

\author{Daisy Zhe Wang}
\email{daisyw@ufl.edu}
\affiliation{University of Florida}

\begin{abstract}

Over the past few years, large knowledge bases have been constructed to store
massive amounts of knowledge. However, these knowledge bases are highly
incomplete.
To solve this problem, we propose a web-based question answering system
system with multimodal fusion of unstructured and structured information, to 
fill in missing information for knowledge bases.

To utilize unstructured information from the Web for knowledge base completion, 
we design a web-based question answering system using multimodal features and question templates to
extract missing facts, which can achieve good performance with very
few questions. To help improve extraction quality, the question answering system
employs structured information from knowledge bases, such as entity types and
entity-to-entity relatedness.


\end{abstract}


\maketitle

\section{Introduction}

A knowledge base (KB) is usually a data store of structured information about
entities, relations and facts.
We implement web-based question answering (WebQA) to extract missing facts from
textual snippets retrieved from the Web.
We design novel multimodal features and an effective question template selection algorithm for WebQA,
which can achieve better performance with fewer questions than previous work \cite{west2014knowledge}.
We implement the query-driven snippet filtering component in WebQA, which can
greatly reduce the number of snippets for processing on-the-fly and improve the
efficiency of the WebQA pipeline.

The WebQA system first transforms KBC queries to natural language questions and
extracts candidate answers from textual snippets searched by these questions on the Web.
It then maps candidate answers to entities in KBs, utilizes entity category information
and relation schema to filter out incorrect candidate answers.
We exploit unstructured textual snippets and structured information in knowledge
bases such as entity-to-entity relatedness and entity descriptions inside KBs to
extract multimodal features for answer ranking.

To improve WebQA efficiency, we need to reduce the number of questions and the
number of snippets to process for each KBC query.
We propose a greedy question template selection algorithm to select a small set
of question templates with highest KBC performance for each relation.
We conduct query-driven snippet filtering to greatly reduce the number of snippets
to be processed.
With fewer questions and snippets for each KBC query, WebQA can provide fast
responses to user queries.

Our contributions are shown below:
\begin{itemize}
	\item 
	We design and implement a web-based question answering (WebQA) system to
	extract missing facts from the unstructured Web with effective multimodal
	features and question template selection, which can achieve better performance
	with fewer questions than previous work \cite{west2014knowledge}.
	\item 
	To improve efficiency, we employ a few query-driven techniques for web-based
	question answering to reduce the
	runtime and provide fast responses to user queries.
	\item 
	Extensive experiments have been conducted to demonstrate the effectiveness and
	efficiency of our system.
\end{itemize}

\noindent \textbf{Overview} Related work on question answering and multimodal fusion
is introduced and discussed in Section 2.
How we design and implement the web-based question answering system are explained in Section 3.
We demonstrate the effectiveness and efficiency of our system through extensive experiments
in Section 4.
The conclusions and future work of our WebQA system are discussed in Section 5.

\section{Related Work}

Knowledge bases are used for various applications, such as entity linking and entity disambiguation in NLP \cite{nia2014streaming,niauniversity}. 
In this section, we briefly discuss related work on question answering and multimodal fusion.

\subsection{Question Answering}

Open-domain question answering (QA) has been popular for a long time.
QA returns exact answers to natural language questions posed by users.
Since 1999, a specialized track related to QA has been introduced into the annual
competition held at the Text Retrieval Conference \cite{voorhees1999trec}.
Web-based QA systems are highly scalable and are among the top performing systems
in TREC-10 \cite{brill2001data}.
Such systems issue simple reformulations of the queries as questions to a search engine,
and rank the repeatedly occurring N-grams in the top snippets as answers based on
named entity recognition (NER) and heuristic answer type checking.

In our system, we choose web-based question answering to extract missing facts
from unstructured textual snippets for knowledge base completion because of its scalability,
flexibility and effectiveness based on the massive information available on the Web.
Our main focus is not developing better general QA systems, but rather addressing
the issue of how to use and adapt such systems for knowledge base completion.
In \cite{west2014knowledge}, West et al. proposed using question templates
based on relevant information of entities to transform KBC queries to natural
language questions and utilized in-house question answering systems to search
for answers of these questions from the Web.
Compared to \cite{west2014knowledge}, we design our own question templates and
a novel template selection algorithm which can greatly reduce the number of questions
and achieve high performance.

\subsection{Multimodal Fusion}
Multimodal fusion techniques have been widely used for tasks such as information retrieval, information extraction and classification tasks \cite{peng2015probabilistic, peng2016scalable, peng2016multimodal, peng2017multimodal, acm2017web}.
Multimodal fusion can utilize information from multiple types of data sources and employ the correlative and complementary relationship between different modalities \cite{peng2016multimodal}, to achieve higher performance than any single modality approaches in most cases.
In the paper, we fuse unstructured and structured data to improve knowledge base completion quality in our WebQA system.

\section{Web-Based Question Answering}

\begin{figure*}[t]
  	\centering
    \includegraphics[width=1.05\textwidth]{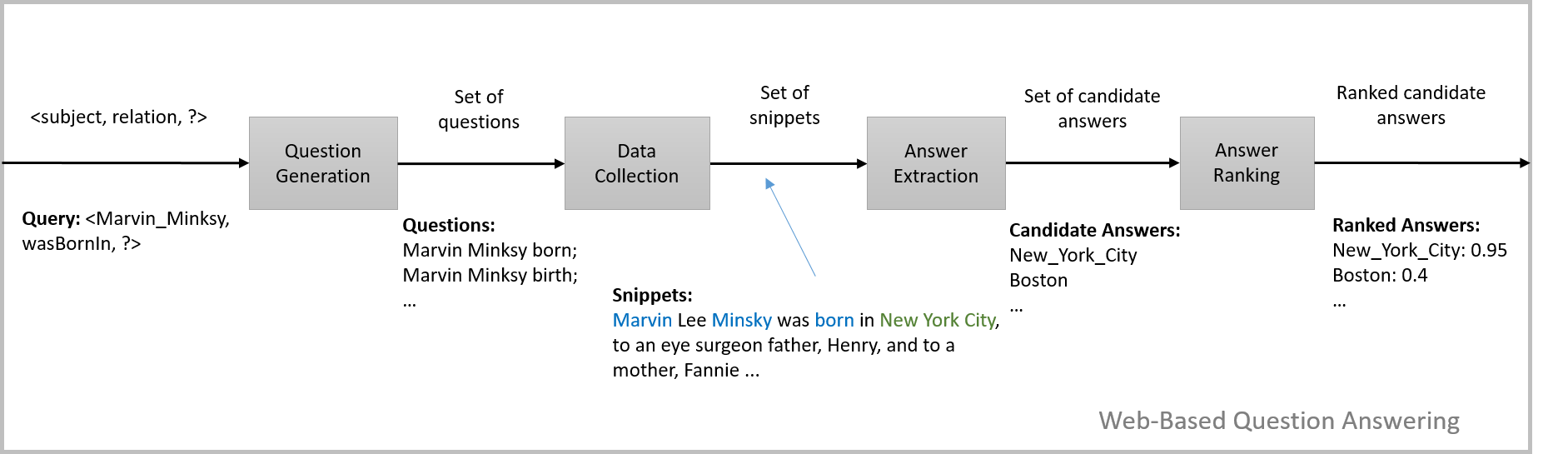}
    \caption{The web-based question answering system pipeline.}
    \label{f2}
\end{figure*}

In this section, we explain the web-based question answering system (WebQA) for
knowledge base completion by fusing both unstructured data from the Web and
structured data from knowledge bases.
WebQA uses question templates to generate multiple natural language questions for each KBC query.
Then textual snippets are crawled by searching these questions on the Web via search engines.
Different from traditional question answering systems, we use entity linking to
collect candidate answers from snippets.
Various multimodal features are extracted for candidate answers by fusing information
from both the unstructured snippets and structured knowledge in KBs.
Then we rank the candidate answers by probability scores generated from classification
on their features.

Compared to previous work \cite{west2014knowledge}, we design better question
templates to achieve high KBC performance.
We propose a greedy algorithm to effectively select a small set of best question
templates for question generation.
We design effective multimodal features through fusion of unstructured textual
snippets and structured knowledge bases.
We conduct query-driven snippet filtering to reduce the number of snippets for
processing, which greatly improves the efficiency of WebQA.
While previous work used batch-oriented question answering systems \cite{west2014knowledge,sun2015open},
WebQA can provide fast responses to user queries.
Experimental results in Section 4 demonstrate both the effectiveness and efficiency of WebQA.

\subsection{WebQA Pipeline}

There are four major components in the WebQA system pipeline, including question generation,
data collection, answer extraction and answer ranking.
The system pipeline of WebQA is illustrated in Figure~\ref{f2}.
We briefly explain the design and implementation of these components.
We use {\textless Marvin\_Minsky, wasBornIn, ?\textgreater} as an example query
and the correct answer to this query is New\_York\_City.
More examples for different relations are shown in Table~\ref{t1}.

\subsubsection{Question Generation}

Structured queries are transformed into natural language questions using selected
question templates, as shown in Table~\ref{t1}.
Each relation has multiple corresponding question templates.
For example, for relation wasBornIn, we use \textit{born}, \textit{birth} and
\textit{birthplace} as its templates.
Then for the KBC query {\textless Marvin\_Minsky, wasBornIn, ?\textgreater},
the corresponding questions are {``Marvin Minsky born"}, {``Marvin Minsky birth"}
and {``Marvin Minsky birthplace"}.
The benefit of using multiple question templates is it can increase the chance
of finding true answers by crawling more snippets with different questions than
using only one template.
As demonstrated by experiments, multiple questions can provide higher KBC performance
than any single question.

In previous work \cite{west2014knowledge}, West et al. utilized relevant information
about entities to augment the questions to design question templates.
For example, for query {\textless Frank\_Zappa, mother, ?\textgreater}, an example
question generated by their templates is {``Frank Zappa mother Baltimore"},
with {Baltimore} being the birthplace of {Frank\_Zappa}.
However, these complex templates tend to generate long questions, which may
return many noisy snippets without true answers by search engines.
For question {``Frank Zappa mother Baltimore"}, search engines may find it hard
to determine whether this question is asking about {``Frank Zappa mother"} or
{``Frank Zappa Baltimore"}, and then return snippets related to {Frank\_Zappa}
and {Baltimore} instead of the mother of {Frank\_Zappa}.

On the contrary, search engines are better at finding relevant snippets to short questions.
Based on this observation, we design question templates by selecting single words
with their meanings close to the semantic meanings of relations.
For example, for relation {wasBornIn}, \textit{born}, \textit{birthplace} and
\textit{birth} are selected as templates; for relation {isMarriedTo},
single words such as \textit{marriage}, \textit{married} and \textit{spouse} are
selected as templates.
More examples about question templates are listed in Table~\ref{t1}.

Issuing all possible questions to search engines is problematic in terms of
computational cost and KBC performance.
To solve this problem, we propose a greedy template selection algorithm to select
a small subset of question templates which achieves the highest KBC performance
for each relation.
The algorithm is explained in Section~\ref{ts}.

\subsubsection{Data Collection}

\begin{table*}[t]
  \caption{Example queries, questions and snippets.} \label{t1}
  \centering
  \small
  \begin{tabular}{lll}
    \hline \hline
    \textbf{Query} & \textbf{Questions} & \textbf{Top snippets} \\
    \hline \hline
    \parbox[t]{5cm}{{\textless Marvin\_Minsky, wasBornIn, ?\textgreater} \newline Templates: born, birth, birthplace, etc.}
    & \parbox[t]{4cm}{ \textit{Marvin Minsky born},\\ \textit{Marvin Minsky birth}, \\ \textit{Marvin Minsky birthplace, etc.}}
    & \parbox[t]{8.5cm}{\textbf{Marvin} Lee \textbf{Minsky} was \textbf{born} in New York City, to an eye surgeon father, Henry, and to a mother, Fannie ... \newline
    \textbf{Marvin Minsky} - A.M. Turing Award Winner, \textbf{BIRTH}: New York City, August 9, 1927. DEATH: Boston, January 24, 2016 ... }\\
    \hline \hline
    \parbox[t]{5cm}{{\textless Ryan\_Block, isMarriedTo, ?\textgreater}  \newline Templates: spouse, married, marriage, etc.}
    & \parbox[t]{4cm}{ \textit{Ryan Block married},\\ \textit{Ryan Block marriage}, \\ \textit{Ryan Block spouse, etc.}}
    & \parbox[t]{8.5cm}{Jul 15, 2014 ... \textbf{Ryan Block}, formerly of Engadget and now at AOL .... More famous for being \textbf{married} to Veronica Belmont IMHO ... \newline
    \textbf{Spouse}(s), Veronica Belmont. \textbf{Ryan Block} (born June 25, 1982) is a San Francisco-based technology entrepreneur ...}\\
    \hline \hline
    \parbox[t]{5cm}{{\textless Julia\_Foster, hasChild, ?\textgreater} \newline Templates: child, children, kid, etc.}
    &\parbox[t]{4cm}{\textit{Julia Foster child},\\ \textit{Julia Foster children,} \\ \textit{Julia Foster kid, etc.}}
    & \parbox[t]{8.5cm}{Mother Love - Ben Fogle and his mother \textbf{Julia Foster} ... A shy and introverted \textbf{child}, he often felt overwhelmed ... \newline
    \textbf{Children}, Ben Fogle, Emily and Bill. \textbf{Julia Foster} (born 2 August 1943) is an English stage, screen and television actress. Born in ...} \\
    \hline \hline
    \parbox[t]{5cm}{{\textless Ruth\_Dyson, isCitizenOf, ?\textgreater} \newline Templates: citizenship, nationality, country, etc.}
    &\parbox[t]{4cm}{\textit{Ruth Dyson citizenship},\\ \textit{Ruth Dyson nationality}, \\ \textit{Ruth Dyson country, etc.}}
    &\parbox[t]{8.5cm}{\textbf{Nationality}, New Zealand. Political party, Labour Party ... \textbf{Ruth} Suzanne \textbf{Dyson} (born 11 August 1957) is a New Zealand politician ... \newline
    \textbf{Ruth} Suzanne \textbf{Dyson} (born 11 August 1957) is a New Zealand politician ... so Dyson's family frequently moved around the \textbf{country}. } \\
    \hline \hline
  \end{tabular}
\end{table*}

We search the generated questions on the Web via search engines and process the
snippets returned by search engines to extract missing facts for KBC queries.
A snippet is a small piece or fragment of text excerpted from the document which
search engines find relevant to the queries.
For query {\textless Marvin\_Minsky, wasBornIn, ?\textgreater}, a top snippet
we crawled from the Web is {"{Marvin} Lee {Minsky} was {born} in New York City,
to an eye surgeon father, Henry, and to a mother, Fannie ...,"} which contains the
correct answer New\_York\_City.
Examples of top snippets for more KBC queries are shown in Table~\ref{t1}.

We crawl up to 50 snippets for each question and hundreds of snippets for each KBC query.
To reduce time waiting for responses from search engines for each relation,
multithreading is employed to parallelize the snippet crawling step with multiple questions.
However, entity linking on hundreds of snippets is still very time-consuming
and far from being able to provide fast responses to user queries.
Therefore, we implement a query-driven snippet filtering component to automatically
select best snippets to extract candidate answers for knowledge base completion,
which is explained in Section~\ref{sf}.

\subsubsection{Answer Extraction}

Noun phrases are extracted from the snippets and linked to entities in knowledge bases.
These linked entities are treated as candidate answers for corresponding questions.
Entity linking is the task to link entity mentions in text with their corresponding
entities in a knowledge base \cite{shen2015entity}.
Linking candidate answers in snippets to entities in knowledge bases has several
remarkable advantages \cite{sun2015open}.
First, redundancy among candidate answers is automatically reduced.
Second, the types of a candidate answer can be effortlessly determined by its
corresponding entity in knowledge bases.
Third, we can develop semantic features for candidate answer ranking by utilizing
the rich semantic information about entities in knowledge bases.

Since entity linking is beyond the scope of this paper, please refer to a survey
paper \cite{shen2015entity} for more information.
An open-source entity linking tool, TagMe \cite{tagme,ferragina2012fast} is employed
in our system to accomplish the entity linking task.
We parallelize the entity linking process using multi-threading to reduce runtime
waiting for responses from a TagMe server \cite{tagme}.

After entity linking, candidate answers with incorrect entity types for KBC queries are discarded.
For example, the query {\textless Marvin\_Minsky, wasBornIn, ?\textgreater} is looking
for candidate answers with type \textit{city} rather than \textit{person}.
In the snippet {"{Marvin} Lee {Minsky} was {born} in New York City, to an eye surgeon
father, Henry, and to a mother, Fannie ..."}, the entity {Henry\_Minsky}, which
is the father of {Marvin\_Minsky}, is discarded because of wrong entity types.
This type filtering step greatly reduces the number of candidate answers for
ranking and thus helps improve answer ranking quality.

\subsubsection{Answer Ranking}

After obtaining a set of eligible candidate answers with correct entity types
from snippets, we extract features for candidate answers from snippets and knowledge bases
and apply classification on these features of candidate answers for ranking.
For feature extraction, we design multimodal features to combine information from
unstructured snippets and structured knowledge in knowledge bases.
The probability scores from classification results are used to rank the candidate answers.

\paragraph{Feature Extraction}

For feature extraction, we adopt the early fusion scheme, which combines information
from multiple modalities at the feature level \cite{peng2015probabilistic,peng2016multimodal,atrey2010multimodal}.
Both unstructured textual snippets from the Web and structured knowledge from KBs
are fused together to produce various effective features in our system.

For each candidate answer, we extract 4 features as shown below.
\begin{itemize}
\item The feature \textit{snippet count} represents the number of snippets in
which a candidate answer appears.
\item The feature \textit{average rank} calculates the average rank of the snippets
in which the candidate answer appears.
\item The feature \textit{average distance} is average distance (number of words)
between the candidate answer and the subject entity in the snippets.
\item The feature \textit{relatedness}\footnote{The entity relatedness implementation is
provided by TagMe \cite{tagme}.} between the candidate answer and the subject entity
measures the semantic relevance of these two entities in knowledge bases.
\end{itemize}

As shown above, these features combine information from both unstructured textual
snippets and structured knowledge bases.
The major advantage of applying multimodal fusion at the feature level is that
multimodal features can provide more information than using only textual snippets or knowledge bases.

\paragraph{Classification and Ranking}

Classification on feature vectors of candidate answers is challenging because of
the highly imbalanced training datasets.
The training datasets usually contain 30+ times more negative samples than positive
samples, making the training datasets extremely biased.
We employed resampling to solve the issue of imbalanced training datasets.
The resampling approach samples the existing training datasets to create new
balanced datasets with equal numbers of positive samples and negative samples.
After using resampling, we usually get classifiers with 20\% to 40\% larger PRC
(area under precision-recall curve) than regular classifiers.

Three classification methods have been tested in our system,
logistic regression \cite{west2014knowledge}, decision tree \cite{sun2015open}
and support vector machines.
Through extensive experiments, logistic regression usually performs better
than the other two classifiers for most relations.

\subsection{Query-Driven Optimization}

We apply two query-driven techniques to reduce the number of questions and the
number of snippets processed by WebQA for each KBC query, which can help WebQA
achieve good KBC performance and efficiency.
The models used in template selection and snippet filtering are trained offline.

\subsubsection{Template Selection} \label{ts}

Issuing all possible questions to search engines is problematic for two reasons.
First, its computational cost is too high.
Processing each question involves significant computational resources
(CPU time and web searches) and requires a lot of time waiting for responses
from search engines.
Moreover, more questions return more snippets and entity linking on a large number
of snippets is also very time-consuming.
Second, the KBC performance may deteriorate with more questions.
Not all questions are equally good.
So by asking all possible questions, we are likely to get more false answers,
which affects the performance of answer ranking.

According to previous work \cite{west2014knowledge}, greedy selection is the
best selection strategy among a few selection strategies, including random selection
and sampling-based selection.
In \cite{west2014knowledge}, West et al. first evaluated the KBC performance of
each question template and then greedily selected the top-performing question templates.
However, their algorithm ignored the correlation between templates.
We observe that some top-performing question templates produce mostly overlapping results.
So we propose a more complex greedy algorithm to learn the best set of question
templates as shown in Algorithm~\ref{a1}.

Let's say there are three templates in descending order of individual performance,
$t_1$, $t_2$ and $t_3$.
Previous work \cite{west2014knowledge} greedily selects top 1, top 2 and top 3 templates,
producing three sets of templates $\{t_1\}$, $\{t_1, t_2\}$ and $\{t_1, t_2, t_3\}$ and
chooses the set of templates with highest performance.
However, $t_1$ and $t_2$ may produce mostly overlapping results.
So $\{t_1, t_2\}$ may not achieve as good KBC performance as $\{t_1, t_3\}$.
Our algorithm also starts with the set $\{t_1\}$.
Then we try to add a new template to the set $\{t_1\}$ to get a larger set with
2 templates.
We compare the performance of $\{t_1, t_2\}$ and $\{t_1, t_3\}$
(all possible size-2 sets with $t_1$ in them) and choose the better set.
The algorithm goes on to add more templates.
Finally we choose the set of templates which achieves the best performance with smallest size.
The details of the algorithm is shown in Algorithm~\ref{a1}.

\begin{algorithm}
\caption{Greedy selection algorithm}
\label{a1}
\begin{algorithmic}[1]
    \STATE $T = \{t_1, t_2, ..., t_n\}$: the set of n question templates\;
    \STATE $Q = \emptyset$: current set of selected question templates\;
    \STATE $QS = \emptyset$: the set of question template sets with different sizes\;
    \FOR{$i = 1$; $i <= n$; $i{+}{+}$}
        \STATE Select $t_j$ from $T$ such that $Q \cup \{t_j\}$ has the highest
        performance for all possible $t$ in $T$\;
        \STATE $Q = Q \cup \{t_j\}$\;
        \STATE $QS = QS \cup \{Q\}$\;
        \STATE $T = T - \{t_j\}$\;
    \ENDFOR
    \STATE Select $Q_m$ from $QS$ with the highest performance and smallest size\;
    \RETURN $Q_m$\;
\end{algorithmic}
\end{algorithm}

The advantage of our greedy selection algorithm is by choosing templates which
work best together, we can avoid computing the exponential combinations of question
templates and solve the correlation problem between top-performing templates to
quickly find the optimal set of question templates.
As shown in experiments, our system can achieve quite good performance with two
or three question templates compared to using all question templates.

\subsubsection{Snippet Filtering} \label{sf}

To reduce the number of snippets for processing, we propose a query-driven
snippet filtering algorithm to select snippets most likely containing information
relevant to knowledge base completion queries.
An important observation is not all top snippets ranked by search engines contain
useful information for KBC queries.
For example, for question {``Marvin Minsky born"}, some of the top snippets returned
by search engines focus on general information about {Marvin\_Minsky} rather than
the birthplace of him.
To solve this problem, we rerank the snippets by classification on features of
them and select top snippets in the reranked list for candidate answer extraction
and ranking.

The features we used for classification on snippets are:
\begin{itemize}
    \item The original rank of a snippet returned by a search engine.
    \item A boolean indicator about whether the question template keyword
     appearing in the snippet or not, e.g. whether \textit{born} appearing in the
     snippets returned by searching the question {``Marvin Minsky born"}.
    \item How many words of entity names appearing in the snippet. For instance,
    if {``Marvin"} and {``Minsky"} both appear inside a snippet for question
    {``Marvin Minsky born"}, the value of this feature is $2$.
\end{itemize}
Clearly these features are designed to select snippets, which not only are originally
high-ranking snippets returned by search engines, but also contain information about
question template keywords and subject entities.

A logistic regression classifier is trained on training datasets and used to filter
snippets on-the-fly in the data collection step of the WebQA pipeline.
The confidence scores of these snippets generated by the classifier are used for snippet reranking.
The original training dataset is also highly imbalanced with positive samples
much fewer than negative samples.
We resolve this issue by conducting resampling on these biased datasets to
generate a new balanced training dataset.

\section{Experimental Results}

In this section, we demonstrate the effectiveness and efficiency of our system
through extensive experiments.
We choose Yago as the knowledge base for its popularity in research community,
its rich ontology and large amount of facts.
We choose four relations (wasBornIn, isMarriedTo, hasChild, isCitizenOf), which
are popular relations frequently studied in previous work.

For KBC performance, we evaluate the quality of candidate answer rankings using
mean average precision (MAP).
For a KBC query, the average precision is defined as
$AP = (\sum_{k=1}^{n} p(k) \times r(k)) / n $,
where $k$ is the rank in the sequence of candidate answers,
$n$ is the number of candidate answers, $p(k)$ is the precision at cut-off $k$
in the ranked list and $r(k)$ is the change in recall from candidate answers $k-1$ to $k$.
Averaging over all queries yields the mean average precision (MAP).

\subsection{Datasets}

Yago \cite{yago,suchanek2007yago} is a huge semantic knowledge base, derived from
Wikipedia, WordNet and GeoNames.
Currently, Yago has knowledge of more than 10 million entities
(like persons, organizations, cities, etc.) and contains more than 120 million
facts about these entities.
The whole Yago knowledge base can be downloaded from Yago website
\footnote{\url{http://www.mpi-inf.mpg.de/departments/databases-and-information-systems/research/yago-naga/yago/downloads/}.}.

We consider 4 relations from Yago (hasChild, isCitizenOf, isMarriedTo and wasBornIn) for
evaluating our system.
To collect training and testing data, we make the local closed-world assumption,
which assumes if Yago has a non-empty set of objects $O$ for a given subject-relation pair,
then $O$ contains all the ground-truth objects for this subject-relation pair.
For each relation, we randomly sampled 500 queries (subjects and corresponding objects) from
Yago as training datasets to train WebQA and 100 queries for testing.

\subsection{KBC Performance}

We first discuss the performance of the WebQA system and different approaches optimizing the
performance of WebQA.
Then we explain the runtime efficiency of our system and how we provide fast responses
to queries on-the-fly.

Logistic regression was chosen as the classification method for answer ranking in WebQA.
To balance training datasets, we applied resampling to the datasets.
We first show experimental results of WebQA using templates selected by Algorithm~\ref{a1}
for 4 relations.
Then, we evaluate the performance of WebQA with query-driven snippet filtering.

\subsubsection{Question Template Selection}

\begin{figure}[t]
  \centering
  \includegraphics[width=0.52\textwidth]{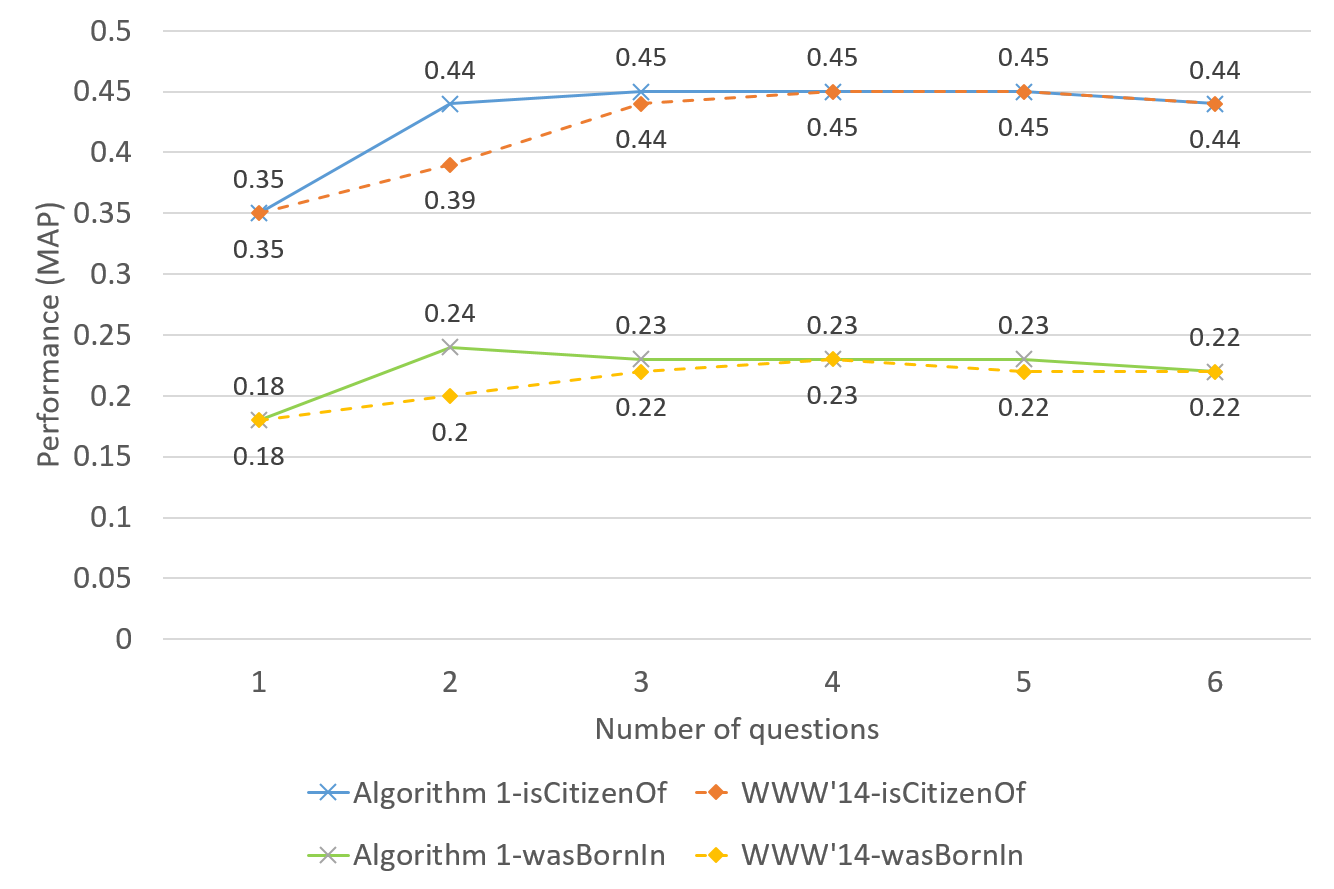}
  \caption{The KBC performance results comparing our greedy selection algorithm (Algorithm 1) with the algorithm in previous work (WWW'14).}
  \label{fgs}
\end{figure}

We compare the performance of Algorithm~\ref{a1} with the greedy algorithm described
in previous work \cite{west2014knowledge}.
Since West et al. didn't enclose the full set of question templates and the question
answering system they used in \cite{west2014knowledge}, we implemented their algorithm
with our templates and the WebQA pipeline.
The results comparing these two algorithms for two relations (isCitizenOf and wasBornIn)
are shown in Figure~\ref{fgs}.

In Figure~\ref{fgs}, we can see multiple questions have higher MAP than using
only one question, although it is generated by the best template.
Another important observation from the results is the KBC performance may deteriorate
using more questions.

For both relations, our greedy algorithm can learn the best set of templates with
fewer templates than the algorithm in WWW'14 \cite{west2014knowledge}.
For relation isCitizenOf, the two algorithm achieve the same highest performance
with different numbers of questions.
For relation hasChild, the best performance of our algorithm is higher than the algorithm in WWW'14.
Other relations demonstrate similar results as isCitizenOf and hasChild.

To conclude, the experimental results in Figure~\ref{fgs} demonstrate our greedy algorithm can
effectively select very few question templates to achieve high KBC performance.
With only two or three questions selected by Algorithm~\ref{a1}, we can achieve
the highest KBC performance.
Hence we can improve the efficiency of the WebQA pipeline with fewer questions
and fewer snippets crawled from them.

\subsubsection{Overall System Performance}
\begin{table*}[t]
  \caption{KBC performance of WebQA with template selection measured by MAP.}
  \label{twt}
  \centering
  \small
  \begin{tabular}{c|cc|cc}
    \hline
    \textbf{Relation} & \textbf{Perf. (WebQA)} & \textbf{Question \# (WebQA)} & \textbf{Perf. (WWW'14)} & \textbf{Question \# (WWW'14)} \\
    \hline
    wasBornIn &  \textbf{0.75} & \textbf{2} & 0.67 & 8  \\
    \hline
    hasChild & \textbf{0.24} & \textbf{2}  & 0.18 & 8 \\
    \hline
    isMarriedTo & \textbf{0.52} & \textbf{3} & 0.50 & 8 \\
    \hline
    isCitizenOf & 0.45 & \textbf{3} & \textbf{0.93} & 32 \\
    \hline
  \end{tabular}
\end{table*}


For each relation, we learned the smallest set of question templates which can achieve
the highest KBC performance using Algorithm~\ref{a1}.
Using these sets of question templates, we conducted experiments to evaluate the overall performance of WebQA for all 4 relations.
These experiments used all snippets crawled from the Web.

In previous work \cite{west2014knowledge}, West et al. designed their own question
templates and employed an in-house question answering system.
They didn't provide their templates or their benchmark datasets \footnote{In previous
work \cite{west2014knowledge}, West et al. conducted experiments based on Freebase.
However, since Freebase is shutdown currently, we prepared our datasets based on Yago.
Freebase and Yago shares similar relations so we can compare them.}.
In their experiments, they evaluated top search entities on Google.com, while we used randomly selected entities.
We compare the KBC performance of WebQA on our benchmarks with their results
for 4 relations, wasBornIn, hasChild, isMarriedTo and isCitizenOf.
The results are shown in Table~\ref{twt}.

For three relations wasBornIn, hasChild and isMarriedTo, our system can achieve
better performance than previous work \cite{west2014knowledge} with much fewer questions,
although we evaluated randomly selected entities.
The performance gain is due to a few reasons.
First, we design better templates and a better template selection algorithm than
previous work, as discussed in Section 4.
Second, we fuse information from both the unstructured text and structured knowledge
bases to design features, while previous work only uses textual information to rank candidate answers.
Only for relation isCitizenOf, our system fails to match previous work.
The possible reason is, previous work \cite{west2014knowledge} evaluated popular entities
on the Web, while the entities we tested are more likely to be rare entities, which don't
have a lot of information for isCitizenOf.
Even though with randomly selected entities, WebQA achieves better performance
with much fewer questions than previous work \cite{west2014knowledge} for wasBornIn, hasChild and isMarriedTo.
So we can expect under the same circumstances, WebQA should achieve (much) better
performance than WWW'14.

\subsubsection{Performance with Snippet Filtering}

\begin{table*}[t]
  \caption{KBC performance of WebQA with snippet filtering for different numbers of snippets. Performance is measured by MAP.}
  \label{tws}
  \centering
  \small
  \begin{tabular}{c|c|c|c|c|c}
    \hline
    \textbf{Relation} & \textbf{10 snippets} & \textbf{20 snippets} & \textbf{30 snippets} & \textbf{All snippets} & \textbf{WWW'14} \\
    \hline
    wasBornIn &  0.70 & 0.71 & 0.70 & 0.75 & 0.67 \\
    \hline
    hasChild & 0.21 & 0.21 & 0.24 & 0.24 & 0.18 \\
    \hline
    isMarriedTo & 0.48 & 0.50 & 0.51 & 0.52 & 0.50 \\
    \hline
    isCitizenOf & 0.39 & 0.40 & 0.41 & 0.45 & 0.93 \\
    \hline
  \end{tabular}
\end{table*}

To reduce the number of snippets for processing, we apply query-driven snippet
filtering to select useful snippets which most likely contain relevant information to queries.
While improving system efficiency, we want to demonstrate through experiments,
selecting a subset of the snippets by query-driven snippet filtering does not
cause severe loss of answer ranking quality.
So we conducted a few experiments with different numbers of snippets and compare
their KBC performance with experiments using all snippets and previous work \cite{west2014knowledge}.
The results are shown in Table~\ref{tws}.

The performance of WebQA using snippet filtering with 10, 20 or 30 snippets
decreases very little compared to using all snippets, with less than 0.04 loss in MAP.
And for relation hasChild, our system achieves the same MAP with 30 snippets as all snippets.
Compared to previous work \cite{west2014knowledge}, WebQA still achieves better
performance for relation wasBornIn, isMarriedTo and hasChild after using snippet filtering.

\subsection{Efficiency}

In our KBC system, we employ a few query-driven approaches to improve the efficiency of WebQA.
Previous work \cite{west2014knowledge,sun2015open} built batch-oriented QA systems
without query-driven optimization, which cannot provide fast responses to user queries.
They didn't provide runtime results of their systems for our system to compare with,
and they didn't focus on the runtime of a single query.

To evaluate the efficiency of our system, the experiments were run on a single
machine with a 3.1GHZ four-core CPU and 4GB memory.
The system runtime varies with multiple environment factors such as network congestion and server speed.
So we calculated average runtime through extensive experiments with multiple queries.


The bottleneck of WebQA is data collection and answer extraction, which involve
web searches and server inquiries.
First, we need to wait for responses from web servers after issuing queries to them.
Second, when we issue multiple queries in parallel to web servers,
there is a considerable delay at the server side to process all queries.

A sequential WebQA pipeline usually costs a few minutes to finish.
So we employed multithreading to parallelize snippet crawling and entity linking
to reduce the time waiting for responses from search engines and TagMe.
A parallelized pipeline achieves about 10x speedups compared to a sequential pipeline.
However, parallelization alone cannot provide fast responses to user queries,
because too many queries are issued simultaneously, causing long network and server-side delays.

\begin{figure}[ht]
  \centering
  \includegraphics[width=0.47\textwidth]{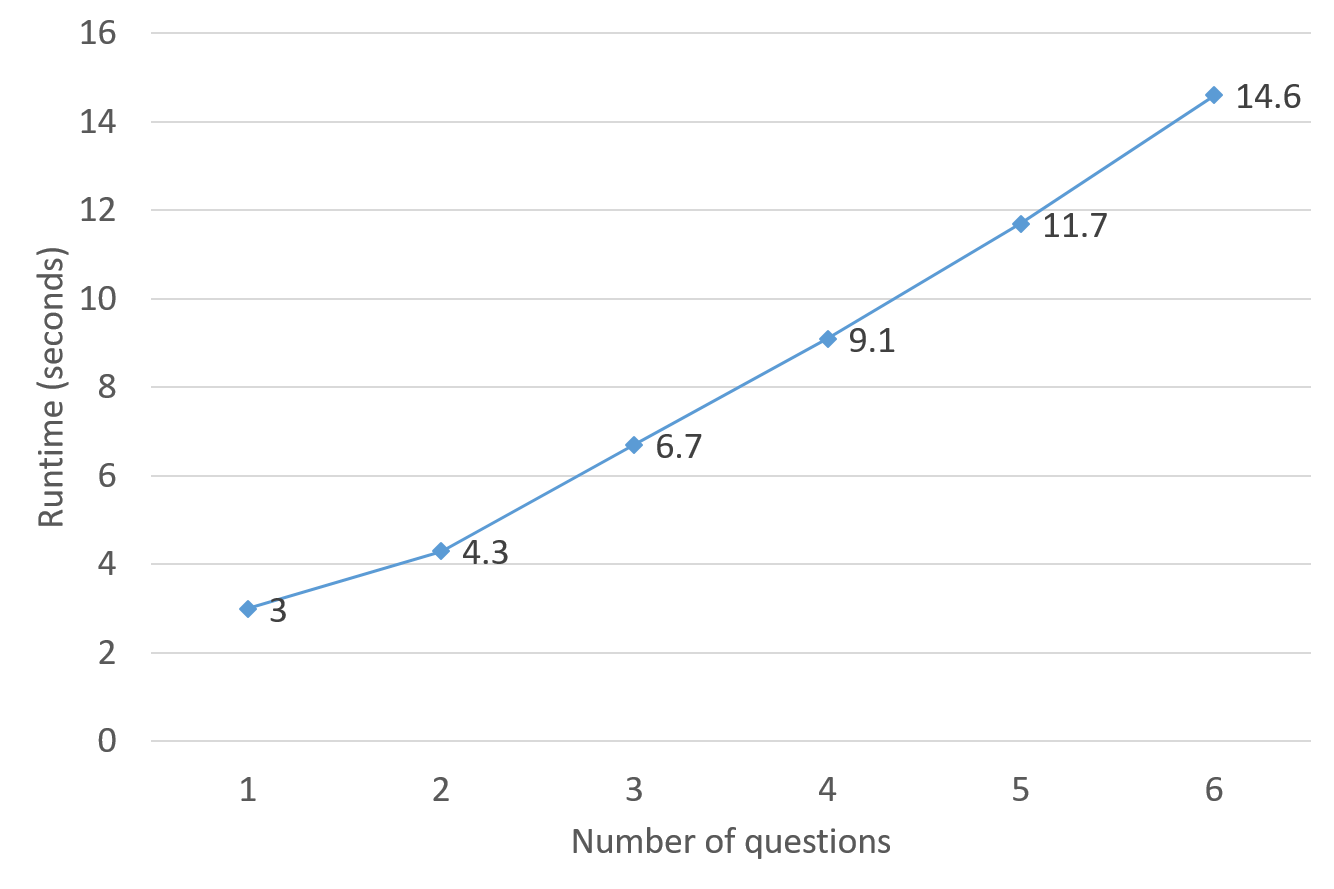}
  \caption{The average runtime of WebQA with different numbers of questions for relation isCitizenOf.}
  \label{fqt}
\end{figure}

\subsubsection{Template Selection}

Experimental results for evaluating the runtime of WebQA with different numbers
of questions for relation isCitizenOf are shown in Figure~\ref{fqt}.
The results for other relations are similar to isCitizenOf.
From Figure~\ref{fqt}, the runtime of WebQA grows almost linearly as the number
of questions increases.
With more questions, WebQA has more snippets to process and entity linking on
a lot of snippets is very time-consuming.
Since our system needs much fewer questions compared to previous work \cite{west2014knowledge},
it is definitely more efficient under the same circumstances.

\subsubsection{Snippet Filtering}
Query-driven snippet filtering is conducted to further improve the runtime by
reducing the number of snippets.
Experimental results of our system using snippet filtering with different numbers
of snippets for relation wasBornIn are shown in Table~\ref{t4}.
Other relations have similar results.
With 3 questions, we need to crawl up to $3 \times 50 = 150$ snippets per query.
Using snippet filtering, we can reduce the number of snippets from $150$ to $20/30$
without too much quality loss as shown in Table~\ref{tws}.
With the number of snippets increasing beyond 30, the server-side delay grows very fast.
The runtime is about $3$ seconds when the number of snippets decreases to $20/30$,
which is about $25\%$ of the time when using all snippets.
In conclusion, WebQA can provide fast responses to user queries, since it only spends
a few seconds for each query.

\begin{table}[ht]
  \caption{Average runtime of WebQA using snippet filtering for relation wasBornIn with 3 questions.}
  \label{t4}
  \centering
  \small
  \begin{tabular}{|c|c|c|c|c|c|c|}
    \hline
    Snippet number & 10 & 20 & 30 & 50 & 100 & 150 \\
    \hline
    Time (seconds) & 2.7 & 3.1 & 3.2 & 4.1 & 7.7 & 12.4 \\
    \hline
  \end{tabular}
\end{table}

\section{Conclusions}

In this paper, we design and implement a web-based question answering (WebQA) system to extract
missing facts from the unstructured Web with effective question templates and
multimodal features, which can achieve better performance with fewer questions
than previous work.
To improve efficiency, we employ a set of query-driven techniques to reduce the
runtime on-the-fly and provide fast responses to user queries.
Extensive experiments have been conducted to demonstrate the effectiveness and
efficiency of our system.
For the future work, we plan to combine both question answering and rule inference via multimodal knowledge graphs
to further improve knowledge base completion quality.


\bibliographystyle{ACM-Reference-Format}
\bibliography{reference}

\end{document}